\documentclass[lettersize,journal]{IEEEtran}
\usepackage{amsmath,amsfonts}
\usepackage{algorithmic}
\usepackage{algorithm}
\usepackage{array}
\usepackage[caption=false,font=normalsize,labelfont=sf,textfont=sf]{subfig}
\usepackage{textcomp}
\usepackage{stfloats}
\usepackage{url}
\usepackage{verbatim}
\usepackage{graphicx}
\usepackage{cite}
\usepackage[symbol]{footmisc}
\usepackage[flushleft]{threeparttable}
\usepackage{xcolor}
\graphicspath{{./Figures/}}

\begin{document}

\title{Real-Time Trajectory Generation and Hybrid Lyapunov-Based Control for Hopping Robots}

\author{Matthew Woodward$^{1}$
\thanks{$^{1} $The author is with the Robot Locomotion and Biomechanics Laboratory.
        {\tt\footnotesize matthew.woodward@tufts.edu}}%
}

\markboth{arXiv. Preprint Version. Submitted October, 2025.}%
{Woodward: Trajectory Generation}


\maketitle

\begin{abstract}
The advent of rotor-based hopping robots has created very capable hopping platforms with high agility and efficiency, and similar controllability, as compared to their purely flying quadrotor counterparts. Advances in robot performance have increased the hopping height to greater than 4 meters and opened up the possibility for more complex aerial trajectories (i.e., behaviors). However, currently hopping robots do not directly control their aerial trajectory or transition to flight, eliminating the efficiency benefits of a hopping system. Here we show a real-time, computationally efficiency, non-linear drag compensated, trajectory generation methodology and accompanying Lyapunov-based controller. The combined system can create and follow complex aerial trajectories from liftoff to touchdown on horizontal and vertical surfaces, while maintaining strict control over the orientation at touchdown. The computational efficiency provides broad applicability across all size scales of hopping robots while maintaining applicability to quadrotors in general.  
\end{abstract}


\section{Introduction}
\IEEEPARstart{H}{opping} robots have shown remarkable efficiency as compared to their flying counterparts \cite{burns_design_2025,burns_optimized_2024,bai_agile_2024,wang_terrestrial_2023}, however both the newer rotor-based and traditional hopping systems \cite{Raibert1984, Raibert1984a} operate in the range of 0.6 to 1.6 meters without significant deviation from a predominantly ballistic trajectory. However, as our pervious work on the MultiMo-MHR showed significant increases in hopping performance ($>4$ m), the aerial phase has sufficient time and energy to begin, as with aerial systems, controlling the overall trajectory between liftoff (LO) and touchdown (TD), allowing for greater agility and adaptability in unstructured terrain. However, unlike aerial systems, trajectory generation for hopping robots must strictly control the TD states to ensure proper positioning, orientation, and foot-surface contact to avoid damage.

To date there exists five untethered continuous hopping robots including: MultiMo-MHR (our robot) \cite{burns_design_2025}, PogoDrone \cite{Zhu2022}, Hopcopter \cite{bai_agile_2024}, Salto/Salto-1P \cite{Haldane2016,haldane_power_2016,Plecnik2017,lee_self-engaging_2018, haldane_repetitive_2017, yim_precision_2018, yim_drift-free_2019,Yim2020}, and PogoX \cite{wang_terrestrial_2023,kang_fast_2024} and one tethered continuous insect-scale hopping robot \cite{hsiao_hybrid_2025}; where, the hopping controllers focus on foot placement and orientation at TD. This allows for the subsequent LO state to be controlled facilitating control over the horizontal locomotion path and stability of the hopping cycle. Hopcopter and Salto have both explored hopping from vertical surfaces (i.e., walls). However, the vertical surface hopping controllers typically control orientation only. The Hopcopter transitions from horizontal flight control to an orientation hold controller at 1.8 m from the wall, and the flight controller is reactivated after the wall-hop. Whereas, Salto initiates a wall-hop from a prior ground-hop oriented towards the wall. At LO an orientation hold controller activates to maintain a prescribed wall contact angle, where presumably the foot placement and orientation hold controller would be reactivated; however this is not discussed. In all cases the trajectories are predominantly ballistic however, to accommodate uncertain LO states and desired TD states, interact with both horizontal and vertical surfaces, and avoid obstacles, control over the entire trajectory from LO to TD is necessary to continue advancing the capabilities of rotor-based hopping robots.

The paper is organized as follows, with Section 2 presenting the dynamic model and the differential flatness derivation. Section 3 develops the real-time hopping trajectory generation methodology, and Section 4 derives the Lyapounv-based controller. Section 5 discusses the trajectory tracking results and Section 6 summarizes the work. 

\begin{figure}[tbp]
\centerline{\includegraphics[width=0.5\textwidth]{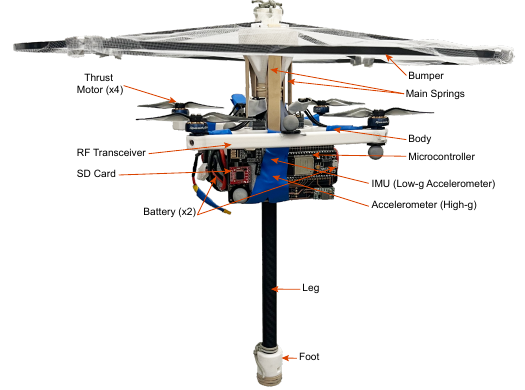}}
\caption{Photo of the MultiMo-MHR with components labeled.}
\label{fig:robot}
\end{figure}

\section{Model}
The current generation of rotor-based hopping robots can be model as a quadrotor during the aerial phase of hopping locomotion with world frame basis $\{\mathbf{x}_W,\mathbf{y}_W,\mathbf{z}_W\}$ and body frame basis $\{\mathbf{x}_B,\mathbf{y}_B,\mathbf{z}_B\}$ both measured in the world frame, and body frame angular velocity $\omega$ measured in the body frame (Fig. \ref{fig:robot}). As with \cite{mellinger_minimum_2011, Bouabdallah2006}, neglecting the rotor dynamics, the Newton-Euler equations of motion about the center-of-mass $\mathbf{r}_{cm}$ are as follows,
\begin{align}
    m_r \ddot{\mathbf{r}}_{cm} &=  - m_r g \mathbf{z}_W + U_1 \mathbf{z}_B - \mathbf{D}_T \label{eq.1}\\ 
    \mathbf{I}_r \dot{\mathbf{\omega}} &= - \mathbf{\omega} \times \mathbf{I}_r \mathbf{\omega} + [U_2, U_3, U_4]^T - \mathbf{D}_R \label{eq.2}
\end{align}
where $m_r$ is the robot mass, $\mathbf{I}_r = \mathrm{diag}[I_x,I_y,I_z]$ is the rotational inertia matrix, $g$ is gravity, $\mathbf{D}_T$ is the translational drag force vector, and $\mathbf{D}_R$ is the rotational drag torque vector. The rotors produce both a force $F_{mi} = \zeta_t \Omega_{mi}$ and torque $\tau_{mi} = \zeta_d \Omega_{mi}$ as a function of their angular velocity $\Omega_{mi}$, thrust factor $\zeta_t$, and drag factor $\zeta_d$. Given the rotor configuration in relation the the body frame of the MultiMo-MHR, the inputs are as follows,
\begin{align}
    \begin{bmatrix}
        U_1 \\
        U_2 \\
        U_3 \\
        U_4
    \end{bmatrix} = 
    \begin{bmatrix}
        \zeta_t & \zeta_t & \zeta_t & \zeta_t \\
        -\zeta_t L_m & -\zeta_t L_m & \zeta_t L_m & \zeta_t L_m \\
        -\zeta_t L_m & \zeta_t L_m & \zeta_t L_m & -\zeta_t L_m \\
        \zeta_d & -\zeta_d & \zeta_d & -\zeta_d \\
    \end{bmatrix}
    \begin{bmatrix}
        \Omega_{m1}^2 \\
        \Omega_{m2}^2 \\
        \Omega_{m3}^2 \\
        \Omega_{m4}^2 \\
    \end{bmatrix} \nonumber
\end{align}
where, $L_m$ is the distance from the center of the rotors to the roll and pitch axes. Equations \ref{eq.1} and \ref{eq.2} can be expanded assuming an orientation parameterized by ZYX Euler angles resulting in the acceleration of the robot as,
\begin{align}
    \ddot{x} &= m_r^{-1}((\cos{\phi} \sin{\theta} \cos{\psi} + \sin{\phi} \sin{\psi}) U_1 + D_x) \label{eq.a1}\\
    \ddot{y} &= m_r^{-1}((\cos{\phi} \sin{\theta} \sin{\psi} - \sin{\phi} \cos{\psi}) U_1 + D_y) \label{eq.a2}\\
    \ddot{z} &= -g + m_r^{-1}(\cos{\phi} \cos{\theta} \, U_1 + D_z) \label{eq.a3}\\
    \dot{p} &= I_x^{-1}(q r (I_y-I_z) + U_2 + D_\phi) \label{eq.a4}\\
    \dot{q} &= I_y^{-1}(p r (I_z-I_x) + U_3 + D_\theta) \label{eq.a5}\\ 
    \dot{r} &= I_z^{-1}(p q (I_x-I_y) + U_4 + D_\psi) \label{eq.a6}
\end{align}
where the non-linear drag forces $\mathbf{D}_T$ and torques $\mathbf{D}_R$ are represented as,
\begin{align}
    \mathbf{D}_T &= \textrm{sign}(\dot{\mathbf{r}}_{cm}) \circ (R_{BW} \mathbf{C}_T)  \dot{\mathbf{r}}_{cm}^2 = [D_x, D_y, D_z]^T  \label{eq.d1}\\
    \mathbf{D}_R &= \textrm{sign}(\mathbf{\omega}) \circ \mathbf{C}_R \mathbf{\omega}^2 = [D_{\phi}, D_{\theta}, D_{\psi}]^T, \label{eq.d2}
\end{align}
and $\circ$ is the Hadamard product. This differs from previous works that have linearized about a nominal operating velocity \cite{svacha_improving_2017, faessler_differential_2018},  as hopping robots inherently must undergo significant changes in velocity for locomotion; e.g., MultiMo-MHR operates across $\pm7$ m/s and increases in performance will only increase the linearization error \cite{burns_design_2025}. Therefore, the general non-linear forms are instead used, where, $\mathbf{R}_{BW}$ is the rotation matrix from body to world frame, and $\mathbf{C}_T$ and $\mathbf{C}_R$ are the overall translational and rotational drag coefficient matrices in the body frame, respectively. The drag coefficient matrices include the air density $\rho$ and effective area $A$, as $\mathbf{C}_{T,R} = 0.5 C_{T,R_{i,j}} \rho A$ . The robot states are therefore defined as $\mathbf{x} = [x,y,z,\dot{x},\dot{y},\dot{z},\phi,\theta,\psi,p,q,r]$.

\subsection{Differential Flatness} \label{s.1}
From \cite{mellinger_minimum_2011, faessler_differential_2018}, the quadrotor model is seen to be differentially flat for four inputs. Therefore, all states and inputs can be calculated from four specifically selected flat outputs and their derivatives. The selected flat outputs include the center-of-mass position $\mathbf{r}_{cm} = [x,y,z]^T$ and the yaw angle $\psi$. Therefore, given a desired trajectory in $\nu = [x,y,z,\psi]$, the desired position, velocity, and acceleration of the 6 degrees-of-freedom including $\mathbf{x}_d = [x,y,z, \dot{x},\dot{y},\dot{z}, \phi,\theta,\psi, p,q,r]$ and $\mathbf{x}_{\ddot{d}} = [\ddot{x},\ddot{y},\ddot{z},\dot{p},\dot{q},\dot{r}]$ can be determined. 

From \cite{mellinger_minimum_2011}, the center-of-mass position, velocity, acceleration, and jerk are determined directly from the flat outputs as $[\mathbf{r}_{cm}, \dot{\mathbf{r}}_{cm}, \ddot{\mathbf{r}}_{cm}, \dddot{\mathbf{r}}_{cm}]$ in the world frame, respectively. The acceleration then determines the orientation $\mathbf{R}_{BW} = [\mathbf{x}_B,\mathbf{y}_B,\mathbf{z}_B]$ by,
\begin{align}
    a_{U_1} &= \ddot{\mathbf{r}}_{cm} + g \, \mathbf{z}_W + \frac{\mathbf{D}_T}{m_r} \nonumber \\
    \mathbf{z}_B &= a_{U_1}/||a_{U_1}||, \,\,
    \mathbf{x}_\psi = [\cos{\psi},\sin{\psi},0]^T  \label{eq.4}\\
    \mathbf{y}_B &= \frac{\mathbf{z}_B \times \mathbf{x}_\psi}{||\mathbf{z}_B \times \mathbf{x}_\psi||}, \, \mathbf{x}_B = \mathbf{z}_B \times \mathbf{y}_B \nonumber
\end{align}
where $\mathbf{x}_\psi$ is the x-axis of the intermediate frame created by a yaw rotation. Taking the derivative of equation \ref{eq.1} results in,
\begin{align}
    m_r & \dddot{\mathbf{r}}_{cm} = \dot{U}_1 \mathbf{z}_B + \mathbf{R}_{BW} \, \mathbf{\omega} \times U_1 \mathbf{z}_B \nonumber\\
    &- \mathrm{sign}(\dot{\mathbf{r}}_{cm} ) \circ {((\dot{\mathbf{R}}_{BW} \mathbf{C}_T) \dot{\mathbf{r}}_{cm}^{\circ 2} + 2 (\mathbf{R}_{BW}  \mathbf{C}_T) \dot{\mathbf{r}}_{cm} } \circ \ddot{\mathbf{r}}_{cm})  \nonumber\\
    &- {\delta }( \dot{\mathbf{r}}_{cm} ) \circ {(2 (\mathbf{R}_{BW} \mathbf{C}_T)  \dot{\mathbf{r}}_{cm}^{\circ 2}  \circ \ddot{\mathbf{r}}_{cm} )}
\end{align}
where, the Dirac delta ($\delta$) component will always be zero, and $\dot{\mathbf{R}}_{BW} = \mathbf{R}_{BW} \, \hat{\mathbf{\omega}}$; where, $\hat{\mathbf{\omega}}$ is the skew symmetric matrix of $\mathbf{\omega}$. Adjusting the center-of-mass jerk ($\,\dddot{\mathbf{r}}_{cm}$) by the drag and eliminating the Dirac delta components results in,
\begin{align}
    m_r \dddot{\mathbf{r}}_{cm}^{\,*}  &= \dot{U}_1 \mathbf{z}_B + \mathbf{R}_{BW} \, \mathbf{\omega} \times U_1 \mathbf{z}_B \quad \mathrm{where,} \nonumber\\
    \dddot{\mathbf{r}}_{cm}^{\,*} &= \dddot{\mathbf{r}}_{cm} + \frac{1}{m_r}\mathrm{sign}(\dot{\mathbf{r}}_{cm} ) \circ  (((\mathbf{R}_{BW} \, \hat{\mathbf{\omega}}) \mathbf{C}_T) \dot{\mathbf{r}}_{cm}^{\circ 2} \nonumber \\
    &+ 2 (\mathbf{R}_{BW}  \mathbf{C}_T) \dot{\mathbf{r}}_{cm}  \circ \ddot{\mathbf{r}}_{cm}). 
\end{align}
Assuming the mass normalized thrust rate of change ($\dot{U}_1/m_r$) is approximately equal to the body z-axis drag adjusted jerk ($\mathbf{z}_B \, \dddot{\mathbf{r}}_{cm}^{\,*}$),  $\dot{U}_1 \sim m_r \mathbf{z}_B \, \dddot{\mathbf{r}}_{cm}^{\,*}$, the drag adjusted jerk ($\dddot{\mathbf{r}}_{cm}^{\,*}$) then determines the body frame angular velocity as,
\begin{align}
    \mathbf{h}_{\omega} &= \mathbf{R}_{BW} \, \mathbf{\omega} \times \mathbf{z}_B = m_r/U_1(\dddot{\mathbf{r}}_{cm}^{\,*} - (\mathbf{z}_B \dddot{\mathbf{r}}_{cm}^{\,*}) \mathbf{z}_B) \label{eq.3}\\
    p &= -\mathbf{h}_{\omega} \cdot \mathbf{y}_B, \quad q = \mathbf{h}_{\omega} \cdot \mathbf{x}_B, \quad r = \dot{\psi} \mathbf{z}_W \cdot \mathbf{z}_B  \label{eq.8}
\end{align}
where equation \ref{eq.3} is the first derivative of equation \ref{eq.1}. Equation \ref{eq.8}, given the $\mathbf{R}_{BW} \, \hat{\mathbf{\omega}}$, results in three equations and three unknowns that can be solved for desired body rotational velocities $[p,q,r]$. In practice this can be simplified by, assuming a constant symmetric drag coefficient $C_T$ resulting in $\dddot{\mathbf{r}}_{cm}^{\,*} = \dddot{\mathbf{r}}_{cm} + \frac{1}{m_r}\mathrm{sign}(\dot{\mathbf{r}}_{cm} ) \circ  (2 C_T \dot{\mathbf{r}}_{cm}  \circ \ddot{\mathbf{r}}_{cm})$, or by using the angular velocities from the previous step $\mathbf{\omega}_{k-1}$.  The angular acceleration and jerk, given the second and third derivatives of equation \ref{eq.1}, can then be found in the same manner as angular velocity.

\section{Hopping Trajectory Generation}
To generate a trajectory, a set of keyframes $\alpha_i(t_i)$ at specific times $t_i$, is necessary to control the entry conditions, progression through, and exit conditions of the generated trajectory. Pervious work in quadrotors has defined the initial and final keyframes $[\alpha_0,\alpha_{m}]$ as the desired flat outputs $\nu^T$; where the sequence of keyframes is connected together using piecewise polynomials, with smooth transitions, that represent the trajectory in each of the four flat outputs \cite{mellinger_minimum_2011}. 

The hopping locomotion cycle is naturally segmented into the stance phase (TD to LO) and the aerial phase (LO to TD). However, due to the high impact forces and torques, and short duration of the stance phase, the aerial phase is the predominant phase for control, and will be the focus of the trajectory generation here. 

Dividing the aerial phase into keyframes, results in an initial at LO $\alpha_{0}$($t_0$) and a final at TD $\alpha_{m}$($t_{m}$), and the potential for intermediate frames added between. The flat outputs of the keyframes must also be expanded to account for the highly dynamic stance phase and necessity to fully control the TD state to avoid damage and ensure proper subsequent LO state. This is achieved by expanding the keyframes to include the desired flat outputs and the first three derivatives, $\alpha_{i} = [\nu, \dot{\nu}, \ddot{\nu}, \dddot{\nu} \,]$ with $i=[0,\ldots,m]$. The full state $\mathbf{x}$ can now be controlled by using the linear acceleration to control the roll and pitch $[\phi,\theta]$ (equation \ref{eq.4}) and jerk to control the angular velocity $[p,q]$ (equations \ref{eq.3}). 

Given the independence of the flat outputs, the trajectory generation problem can be divided into four separate problems, where, $\mathbf{\nu}_{i,j,k}(t)$ represents the value of the $i^{th}$ keyframe, $j^{th}$ flat output, and $k^{th}$ derivative of the flat output at time $t$. Then the trajectory polynomial $\mathbf{\nu}_{j,k}(t)$ that connects the keyframes of the $k^{th}$ derivative of the $j^{th}$ flat output can be modeled as follows,
\begin{align}
    \mathbf{\nu}_{j,k}(t)& = \frac{d^k}{dt^k}(f_n(t)) \, \mathbf{c}_j  \nonumber\\
    &= \frac{d^k}{dt^k}([1,t,t^2, ..., t^n] ) [a_0, a_1, a_2, ..., a_n]_j^T \label{eq.5}
\end{align}
where, $t_0 \leq t \leq t_{m}$, and $n$ is the order of the trajectory polynomial. As with \cite{van_nieuwstadt_realtime_1998}, to solve for the coefficients $\mathbf{c}_j = [a_0, a_1, a_2, ..., a_n]^T_j$, the problem can be setup as a linear algebraic solution $\mathbf{P}_l \mathbf{c}_j = \mathbf{\nu}_j$, with full keyframes (includes all three derivatives), results in,
\begin{align}
    \mathbf{P}_l &= 
    \begin{bmatrix}
        f_n(t_0) \\
        d f_n(t_0)/dt \\
        d^2 f_n(t_0)/dt^2 \\
        d^3 f_n(t_0)/dt^3 \\
        \vdots \\
        f_n(t_{m}) \\
        d f_n(t_{m})/dt \\
        d^2 f_n(t_{m})/dt^2 \\
        d^3 f_n(t_{m})/dt^3 \\
    \end{bmatrix} 
    , 
    \mathbf{c}_j = 
    \begin{bmatrix}
        a_0 \\ 
        a_1 \\ 
        a_2 \\
        \vdots \\
        a_n
    \end{bmatrix} 
    ,  
    \mathbf{\nu}_j = 
    \begin{bmatrix}
        \nu_{1,j,0}(0) \\ 
        \nu_{1,j,1}(0) \\ 
        \nu_{1,j,2}(0) \\
        \nu_{1,j,3}(0) \\
        \vdots \\
        \nu_{{m},j,0}(t_{m}) \\ 
        \nu_{{m},j,1}(t_{m}) \\ 
        \nu_{{m},j,2}(t_{m}) \\
        \nu_{{m},j,3}(t_{m}) \\
    \end{bmatrix} \nonumber\\
    &\qquad \qquad \quad  _{(4(m+1) \times n+1)} \qquad \, _{(n+1 \times 1)} \qquad \qquad \quad _{(4(m+1) \times 1)} \nonumber
\end{align}
where, $\mathbf{P}_l$ represents the polynomial matrix of $f_n$ and its derivatives $d^k /dt^k (f_n)$,  and $\mathbf{\nu}_j$ is the desired flat output vector. The general solution is therefore $\mathbf{c}_j = \mathbf{P}_l^{-1} \mathbf{\nu}_j$ where $\mathbf{c}_j = \mathbf{c}_j^* + \mathbf{N}_l(\mathbf{P}_l) \mathbf{c}_{j_N}$, $\mathbf{c}_j^*$ is the least squares fit coefficients, $\mathbf{N}_l(\mathbf{P}_l)$ is the null space matrix of $\mathbf{P}_l$, and $\mathbf{c}_{j_N}$ is the null space coefficient vector. While $\mathbf{c}_j^*$ can be efficiently calculated, previous work has shown the null space coefficients $\mathbf{c}_{j_N}$ must be calculated through general optimization \cite{van_nieuwstadt_realtime_1998} or the overall coefficients $\mathbf{c}_j$ can be optimized through quadratic programming to minimized the square of the forth derivative of position (snap) \cite{mellinger_minimum_2011}. Optimization, however, requires time and computational power which limits the potential for real time trajectory generation. To allow real-time generation, normalized trajectories can be precalculated and then, in real-time, temporally and spatially scaled \cite{mellinger_minimum_2011}. However, as scaling fundamentally changes the derivatives of the flat outputs, the final keyframe will not maintain the desired derivative values. Moreover, because of the desired derivative values and operation about zero thrust, instead of hover in quadrotors, scaling time or space will fundamentally and potentially significantly alter the required trajectory. Therefore, whereas quadrotors may be less concerned with the derivatives of the flat outputs at the final keyframe, hopping robots must maintain the desired values,  as they dictate the TD orientation, TD energy, and LO state. Therefore, a computationally efficient real-time trajectory generation methodology for hopping robots will be developed.

\begin{table}[tbp!]
\caption{Hopping Trajectories}
\begin{center}
\begin{tabular}{p{0.05\textwidth}|p{0.02\textwidth}p{0.02\textwidth}p{0.02\textwidth}p{0.02\textwidth}|p{0.03\textwidth}|p{0.02\textwidth}p{0.02\textwidth}p{0.02\textwidth}p{0.02\textwidth}}
\hline
& \multicolumn{4}{c|}{Keyframe ($\alpha_0$) LO}  & ($\alpha_1$) & \multicolumn{4}{c}{Keyframe ($\alpha_2$) TD} \\
\textbf{Traj.} & \textbf{$\nu$}& \textbf{$\dot{\nu}$}& \textbf{$\ddot{\nu}$}& \textbf{$\dddot{\nu}$}& \textbf{$\ddot{\nu}$} &
\textbf{$\nu$}& \textbf{$\dot{\nu}$}& \textbf{$\ddot{\nu}$}& \textbf{$\dddot{\nu}$} \\
\hline
\multicolumn{9}{l}{T1 (LO-TD): Ground-Ground, Wall-Ground}\\
\hline
$x,y$ & x& x& x& x& x& o& x& x& x\\
$z$ & x& x& x& x& x& x& x& x& x\\
$\psi$ & x& x& o& o& o& x& x& o& o\\
\hline
\multicolumn{9}{l}{T2 (LO-TD): Ground-Wall, Wall-Wall}\\
\hline
$x,y$ & x& x& x& x& x& x& x& x& x\\
$z$ & x& x& x& x& x& o& x& x& x\\
$\psi$ & x& x& o& o& o& x& x& o& o\\
\hline
\multicolumn{9}{l}{T3 (LO-TD): Ground-State, Wall-State}\\
\hline
$x,y$ & x& x& x& x& x& x& x& x& x\\
$z$ & x& x& x& x& x& x& x& x& x\\
$\psi$ & x& x& o& o& o& x& x& o& o\\
\hline
\end{tabular}
\begin{tablenotes}
\item The (x) indicates desired values, and the (o) indicates free values.
\end{tablenotes}
\label{tab:hop_traj}
\end{center}
\end{table}

\begin{table}[tbp!]
\caption{Hopping Keyframes}
\begin{center}
\begin{tabular}{p{0.04\textwidth}|p{0.12\textwidth}p{0.02\textwidth}|p{0.17\textwidth}p{0.02\textwidth}}
\hline
& \multicolumn{2}{c|}{Keyframe ($\alpha_0$) LO}  & \multicolumn{2}{c}{Keyframe ($\alpha_2$) TD} \\
\textbf{Traj.} & $x,y,z$ & $\psi$ & $x,y,z$ & $\psi$ \\
\hline
$U_{1_{d}}$ & $U_{1_{LO}} = 0.9 m_r g$ & NA& $U_{1_{TD}} = 0.2 m_r g$ & NA \\
$v_{TD_{d}}$ & NA & NA & DV& NA \\
$\mathbf{z}_{B_d}$ & NA & NA & DV& NA \\
\hline
$\nu$ & SE& SE& DV& DV\\
$\dot{\nu}$ & SE& SE& $-v_{TD} \mathbf{z}_{B_d}$& $0$\\
$\ddot{\nu}$ & $U_{1_{LO}} \mathbf{z}_B - g \, \mathbf{z}_W$ & $0$ & $U_{1_{TD}} \mathbf{z}_{B_d} - g \, \mathbf{z}_W - \frac{\mathbf{D}_T}{m_r}$ & $0$\\
$\dddot{\nu}$ & $[0,0,0]^T$ & $0$ & $[0,0,0]^T$ & $0$ \\
\hline
\end{tabular}
\begin{tablenotes}
\item The (SE) indicates state estimation values, (DV) indicates desired values, and (NA) indicates not applicable. Note: the acceleration of the intermediate keyframe $\alpha_1$ is equal to the acceleration of $\alpha_2$.
\end{tablenotes}
\label{tab:hop_keyframes}
\end{center}
\end{table}

To develop a real-time hopping trajectory generation methodology, three keyframes will be used including, an initial at LO $\alpha_0(t_0=0)$, an intermediate near TD $\alpha_1(t_1)$, and a final at TD $\alpha_2(t_2)$; where $t_2 = t_m$ is the total time from LO to TD, and the intermediate keyframe $\alpha_1$ is used to force the robot to the desired TD orientation prior to the TD keyframe $\alpha_2$. Therefore, keyframe $\alpha_1$ will only contain the accelerations and they will be set equal to the desired accelerations in keyframe $\alpha_2$. The order $n$ of the trajectory polynomial $f_n(t)$  will be selected as one less than the total number of desired flat outputs across the the three keyframes $\mathbf{\nu}_j$; where removing specific desired flat outputs, allows the value to vary and can reduce the aggressiveness of the generated trajectory $\mathbf{c}_j^*$. Therefore, $\mathbf{P}_l$ is square, and the symbolic inverse $\mathbf{P}_l^{-1}(t_m)$ is easily precomputed as a function of both  $t_1$ and $t_2$; which allows for an inherently parallel and therefore efficient calculation of the the least squares coefficients $\mathbf{c}_j^*$.

\begin{table*}[tbp!]
\caption{Control Equations}
\begin{center}
{\large
\begin{tabular}{p{0.95\textwidth}}
\hline
\\ [-2ex]
$U_1 = m_r (  ({\ddot{z} }_d +\sigma_1)+\dot{e}_z  k_{p_z} ) +\frac{k_{U_1}  m_r (\dot{e}_z  k_{d_z} +e_z  k_{p_z} )}{2 k_{d_z} }$ \\
$U_2 = \frac{I_x  (k_{d_y}  ({\ddot{y} }_d +\sigma_2)+\dot{e}_y  k_{p_y} +e_p  k_{p_\phi} )}{L_t  k_{d_\phi} } -\frac{I_x  (\frac{D_{\phi } }{I_x }-\dot{p}_d +\frac{r  q  (I_y -I_z )}{I_x })}{L_t  } +\frac{I_x  k_{U_2}  (\dot{e}_y  k_{d_y} +e_p  k_{d_\phi} +e_\phi  k_{p_\phi} +e_y  k_{p_y}  )}{2 L_t  k_{d_\phi} }$ \\
$U_3 = \frac{I_y  (\dot{e}_x  k_{p_x} +e_q  k_{p_\theta} -k_{d_x}  (-{\ddot{x} }_d +\sigma_3))}{L_t  k_{d_\theta} } + \frac{I_y  (\dot{q}_d -\frac{D_{\theta } }{I_y }+\frac{p  r  (I_x -I_z )}{I_y })}{L_t   } +\frac{I_y  k_{U_3}  (\dot{e}_x  k_{d_x} +e_q  k_{d_\theta} +e_\theta  k_{p_\theta} +e_x  k_{p_x}  )}{2 L_t  k_{d_\theta} }$ \\
$U_4 = I_z  \dot{r}_d -D_{\psi } +\frac{I_z  e_r  k_{p_\psi} +\frac{I_z  e_\psi  k_{U_4}  k_{p_\psi} }{2}}{k_{d_\psi} }$ ; where,  
$\sigma_1 =  -\frac{D_x  (s_\phi s_\psi+c_\phi c_\psi s_\theta)}{m_r} +\frac{D_y  (c_\psi s_\phi-c_\phi s_\psi s_\theta)}{m_r} -\frac{c_\phi c_\theta (D_z -g m_r)}{m_r}$\\
$\sigma_2 = \frac{D_x  (c_\phi s_\psi-c_\psi s_\phi s_\theta)}{m_r} -\frac{D_y  (c_\phi c_\psi+s_\phi s_\psi s_\theta)}{m_r} -\frac{c_\theta s_\phi (D_z -g m_r)}{m_r}$ 
, 
$\sigma_3 = \frac{D_x  c_\psi c_\theta+D_y  c_\theta s_\psi - (D_z - g m_r)  s_\theta}{m_r}$ \\
\\ [-2ex]
\hline
\end{tabular}}
\begin{tablenotes}
\item Implementation: Set [$-e_y, -\dot{e}_y$] for proper control direction. The MultiMo-MHR uses the following: [$k_{px},k_{dx},k_{py},k_{dy},k_{pz},k_{dz}$] = [$10,1,10,1,10,1$], [$k_{p \phi},k_{d \phi},k_{p \theta},k_{d\theta},k_{p \psi},k_{d \psi}$] = [$30,1,30,1,30,1$], and [$k_{U1},k_{U2},k_{U3},k_{U4}$] = [$10,80,80,80$], with the remaining parameters from previous work \cite{burns_design_2025, burns_optimized_2024}. Notation: $s_\gamma=\sin(\gamma)$ and $c_\gamma=\cos(\gamma)$
\end{tablenotes}
\label{tab:control}
\end{center}
\end{table*}
 
To add trajectory flexibility, the order $n$ of the trajectory polynomial $f_n(t)$ in $\mathbf{P}_l$ is increased by $n^*$, adding $n^*$ null space basis vectors to the null space matrix $\mathbf{N}_l(\mathbf{P}_l(f_{n+n^*}))_{(n+n^* \times n^*)}$ which can also be symbolically precomputed as functions of $t_1$ and $t_2$. Therefore, the solution is modified to $\mathbf{c}_j = [(\mathbf{c}_j^*)^T, 0_{(1\times n^* )}]^T + \mathbf{N}_l(\mathbf{P}_l(f_{n+n^*})) \, \mathbf{c}_{j_N}$, where, $0_{(1\times n^* )}$ is a zero vector to account for the added polynomial coefficients not included in $\mathbf{c}_j^*$.  To analytically solve for the null space coefficients $\mathbf{c}_{j_N}$, we will use the superposition principle of the null space.  Therefore, substituting the modified solution into equation \ref{eq.5} and solving for the null space coefficients $\mathbf{c}_{j_N}$ results in,
\begin{align}
    &\mathbf{c}_{j_N} = (\mathbf{M}_1^T \mathbf{M}_1)_{(n^* \times n^*)}^{-1} (\mathbf{M}_1^T \mathbf{M}_2)_{(n^* \times n^*)};  \quad \mathrm{where, }\label{eq.6}\\
    &\mathbf{M}_1 = \mathbf{P}_N \,  \mathbf{N}_l(\mathbf{P}_l(f_{n+n^*})), \nonumber\\
   &\mathbf{M}_2 = \mathbf{\nu}_{j,k}(t) -  \mathbf{P}_N \, [(\mathbf{c}_j^*)^T, 0_{(1\times n^* )}]^T, \nonumber \\
   &\mathbf{P}_{N \, (s \times n+n^*)} = \frac{d^k}{dt^k}(f_{n+n^*}(t)), \nonumber
\end{align}
$\mathbf{\nu}_{j,k}(t)$ is now the vector of the $j^{th}$ desired flat outputs and its $k^{th}$ derivatives at $s$ time points, and $\mathbf{P}_N$ is the polynomial matrix of $f_{n+n^*}$ at $s$ time points. The desired flat outputs can specify any derivative  $k \leq n+n^*$ and any number $s$ of time points $t$ such that $t_0\leq t \leq t_m$ with the exception of the desired points specified in $\mathbf{P}_l$ as the null space will always be zero at those points.  As can be seen in equation \ref{eq.6}, both $\mathbf{M}_1^T \mathbf{M}_1$ and $\mathbf{M}_1^T \mathbf{M}_2$ are square with dimension $n^* \times n^*$. Therefore, regardless of the number of desired flat outputs $\mathbf{\nu}_{j,k}(t)$, calculation of the null space coefficients will always be related to the number of null space basis vectors. Given the number of additional desired flat outputs $s$ along the trajectory is less than or equal to $n^*$, the solution is guaranteed to satisfy them, whereas, if $s>n^*$ the result will be the least squares solution. However, as the fit is only using the null space, the desired LO, intermediate, and TD keyframe values will remain unchanged. It is important to note that increasing the polynomial order $n+n^*$ increases trajectory adaptability but also increases the potential state derivative values, and therefore the difficultly in following the trajectory; where, robot characteristics such as thrust-to-weight and torque-to-rotational inertia will determine the upper limit.

\subsection{Hopping Trajectories}
Given the structured nature of the hop cycle, it is possible to identify general hopping trajectory types based on the TD surface and characteristics. Table \ref{tab:hop_traj} shows the six general hopping trajectories combined into three unique sets of desired and free flat outputs in the three hopping keyframes. These include those that TD on horizontal surfaces (T1: Ground-Ground, Wall-Ground) with position $[x,y]$ free, those that TD on vertical surfaces (T2: Ground-Wall, Wall-Wall) with position $[z]$ free, and those that TD at a specified state (T3: Ground-State, Wall-State) with none free; where, in all cases $[\ddot{\psi},\dddot{\psi} \,]$ are free. Table \ref{tab:hop_traj}, shows three unique rows which creates three unique $\mathbf{P}_l$ matrices resulting in three $\mathbf{P}_l^{-1}$ and $\mathbf{N}_l(\mathbf{P}_l(f_{n+n^*}))$ matrices to precompute, as functions of the trajectory time $t_m$; including, $\mathbf{P}_0$ (all desired), $\mathbf{P}_1$ (free TD position), $\mathbf{P}_2$ (free acceleration and jerk). Since the null space basis vectors abide by the superposition principle, precomputing the null spaces $\mathbf{N}_l(\mathbf{P}_l(f_{n+n^*}))$ for more null space basis vectors than necessary, allows for any individual or combination of basis vectors to be used for each trajectory generated. 

In practice, it is observed that $n^*=2$, provides good flexibility in adjusting the generated trajectory. To set the LO keyframe $\alpha_0$, the position, velocity, and orientation (e.g., Euler angles) must be determined from state estimation along with a desired input $U_{1_{LO}}$. Given the fast response time of the motors and probable desired for high thrust at LO, the acceleration, jerk, and yaw states are set as seen in Table \ref{tab:hop_keyframes}.  To set the TD keyframe $\alpha_2$, the desired position, orientation, velocity magnitude $v_{TD}$, and input $U_{1_{TD}}$ must be first determined, and the velocity, acceleration, jerk,and yaw states are then set as seen in Table \ref{tab:hop_keyframes}. The TD thrust $U_{1_{TD}}$ is set to $20\% $ of the body weight to ensure proper orientation for surface contact with the robot's foot, and the velocity vector $-v_{TD} \mathbf{z}_B$ is aligned with the orientation of the robot; i.e., the TD velocity vector is aligned with the body z-axis creating zero moment at TD.

\begin{figure*}[tbp]
\centerline{\includegraphics[width=0.95\textwidth]{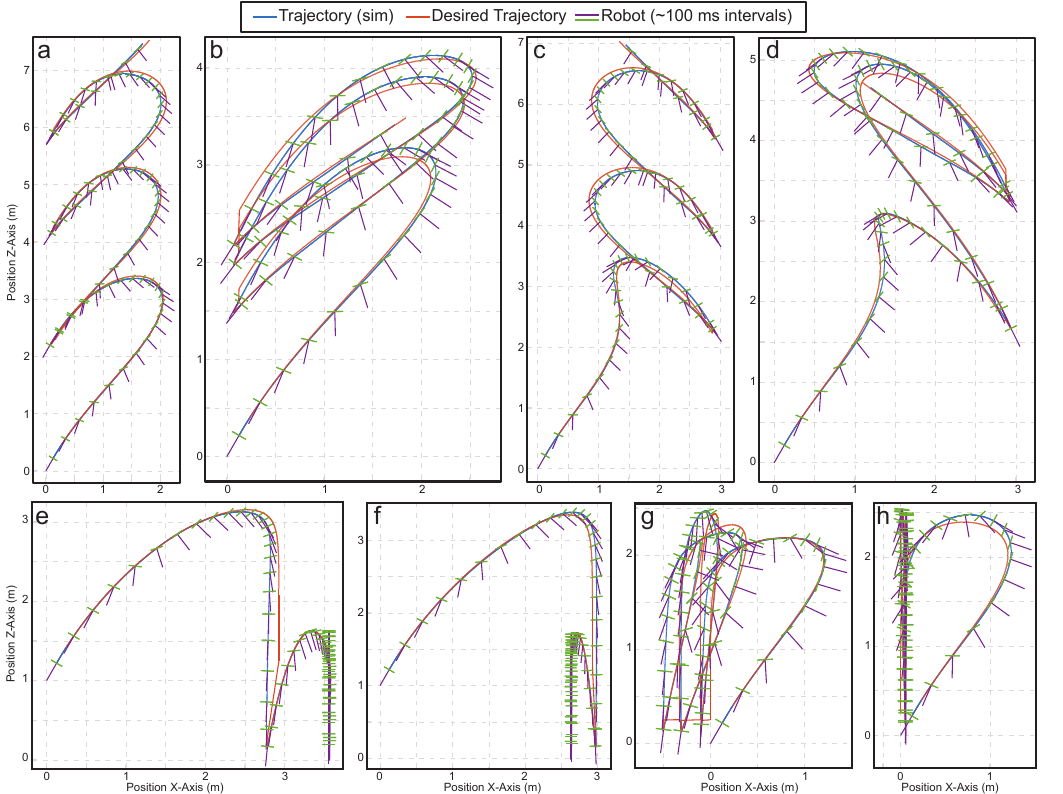}}
\caption{Hopping trajectory generation and control (Tables \ref{tab:hop_keyframes} and \ref{tab:traj_error} show details) for constant initial conditions (keyframe $\alpha_0$) including roll, pitch, and yaw of $[\phi, \theta, \psi] = [0,30,0]$ degrees, velocity magnitude aligned with the body z-axis of $5$ m/s, trajectory time $t_m = 1.75$ seconds (LO-TD), and reorientation time $\delta_t = 0.05$ seconds; where total simulation time equals $6$ seconds. The TD keyframe $\alpha_2$ for all generated trajectories includes $v_{TD_d}=5$ m/s and the desired orientation $[\phi, \theta, \psi]$ is set as: a,b) $[0,30,0]$, c,d) $[0,-30,0]$, e,f) $[0,0,0]$, g,h) $[0,0,0]$ degrees.}
\label{fig:traj_grd_wall}
\end{figure*}

\section{Control}
To follow the generated trajectory, a Lyapunov-based controller will now be developed. First the errors $\mathbf{e} = \mathbf{x}_d - \mathbf{x}$ in position and velocity of all 6-degress-of-freedom (6DOF) are determined, where the translational errors in position and velocity are rotated into the body frame with $\mathbf{R}_{BW}^T$. The errors are then divided into the four control errors as follows,
\begin{align}
    e_{U_1} &= k_{p_z} e_{z} + k_{d_z} \dot{e}_{z} \nonumber\\
    e_{U_2} &= k_{p_y} e_{y} + k_{d_y} \dot{e}_{y} + k_{p_\phi} e_{\phi} + k_{d_\phi} e_{p} \nonumber\\
    e_{U_3} &= k_{p_x} e_{x} + k_{d_x} \dot{e}_{x} + k_{p_\theta} e_{\theta} + k_{d_\theta} e_{q} \nonumber\\
    e_{U_4} &= k_{p_\psi} e_{\psi} + k_{d_\psi} e_{r} \nonumber
\end{align}
where $k_p$'s $\geq 0$ and $k_d$'s $>0$ are the gains. A positive definite Lyapunov candidate function $V(\mathbf{x})$ is then,
\begin{align}
    V(\mathbf{x}) &= \frac{1}{2} \sum_{i=1}^4 e_{U_i}^2
\end{align}
where, the derivative of $V$ is set to be negative definite as,
\begin{align}
    \frac{dV}{dt} &= -\frac{1}{2} \sum_{i=1}^4 k_{U_i} e_{U_i}^2.
\end{align}
The sum in both $V$ and $dV/dt$ allows the terms associated with the four inputs $[U_1,U_2,U_3,U_4]$ to be separated as,
\begin{align}
    \frac{1}{2} \frac{d}{dt} e_{U_i}^2 &= -\frac{1}{2}  k_{U_i} e_{U_i}^2 \label{eq.7}
\end{align}
Computing the derivative will naturally require derivatives of the 6DOF velocities resulting in acceleration errors in equation \ref{eq.7}; e.g., $\ddot{e}_{x} = \ddot{x}_d - \ddot{x}$. To account for the robot dynamics, the translational accelerations (equations \ref{eq.a1}-\ref{eq.a3}), rotated into the body frame by $\mathbf{R}_{BW}^T$, and rotational accelerations (equations \ref{eq.a4}-\ref{eq.a6}) are substituted in for the state accelerations. Table \ref{tab:control} shows the solutions for the individual inputs $U_i$.

The controller represents a hybrid Lyapunov-based controller where the aerial phase (LO-TD) evolves continuously with stability guarantees and the stance phase (TD-LO) is represented as a discrete event. To maintain stability over the complete hop cycle requires maintaining stability guarantees during the stance phase. Therefore, the difference between the candidate functions at TD and the subsequent LO must be as follows,  $V(\mathbf{x}_{LO}) - V(\mathbf{x}_{TD}) \leq 0$; where, the LO states $\mathbf{x}_{LO}$ are determined by a function $h(\mathbf{x}_{TD})$ and the TD states as, $\mathbf{x}_{LO} = h(\mathbf{x}_{TD})$. 

Assuming the robot's desired LO state is equal to the prior TD state with velocity direction changes, and including the stance phase energy losses, $h(\mathbf{x}_{TD})$ does not have to be directly determined. Instead, $V(\mathbf{x}_{LO}) - V(\mathbf{x}_{TD})$ will be a positive constant bound by the magnitude of the TD roll and pitch angles; where, zero desired roll and pitch will result in $V(\mathbf{x}_{LO}) - V(\mathbf{x}_{TD}) \simeq 0$. This is due to the alignment of the velocity vector and the body z-axis at TD, where the only moment during stance is that due to gravity; which tends to increase the roll and pitch angles at LO (Table \ref{tab:hop_keyframes}). This allows for a wide variety of trajectories as seen in Fig. \ref{fig:traj_grd_wall} and \ref{fig:traj_jump}. To stabilize at non-zero roll and pitch angles the desired TD angles can be modified as follows, $[\phi_{TD} \, \gamma_\phi, \theta_{TD} \, \gamma_\theta]$, where the adjustment parameters are learned over multiple hop cycles as,
\begin{align}
    \gamma_{\beta} & = \gamma_{\beta} - \mu \, \mathrm{sign}(\beta_{TD}) \mathrm{sign}(\beta_{LO}) |\beta_{TD} - \beta_{LO}|. \nonumber
\end{align}
The $\beta$ represented the roll and pitch angles $[\phi,\theta]$, $\mu = 0.1$ is the learning rate, and the adjustment parameters are initialized to one, $[\gamma_\phi, \gamma_\theta] = 1$. This drives the LO angles to the TD angles, and therefore, when averaged over multiple hops, leads to overall stability. Finally, to maintain stability when the LO and TD angles differ, the $h(\mathbf{x}_{TD})$ must be determined. This has been achieve through conservation of angular momentum for planar motion \cite{Yim2020} and fitted polynomials to simulated data \cite{yim_precision_2018}. However, a trained neural network could be capable of capturing the characteristic of the real system with the TD velocity and desired LO orientation as inputs, and the required TD orientation as outputs; and will be explored in future work.

\section{Trajectory Tracking Performance}
\begin{figure}[tbp]
\centerline{\includegraphics[width=0.5\textwidth]{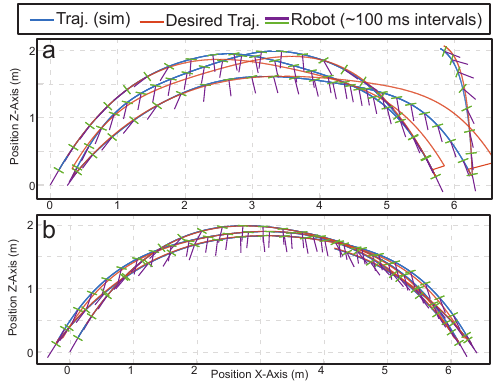}}
\caption{Hopping trajectory generation and control (Tables \ref{tab:hop_keyframes} and \ref{tab:traj_error} show details) for constant initial conditions (keyframe $\alpha_0$) including roll, pitch, and yaw of $[\phi, \theta, \psi] = [0,30,0]$ degrees, velocity magnitude aligned with the body z-axis of $5$ m/s, trajectory time $t_m = 1.75$ seconds (LO-TD), and reorientation time $\delta_t = 0.05$ seconds; where total simulation time equals $6$ seconds. The TD keyframe $\alpha_2$ for all generated trajectories includes $v_{TD_d}=5$ m/s and the desired orientation $[\phi, \theta, \psi]$ alternates between $[0,-30,0]$ and $[0,30,0]$ degrees. a) No drag compensation. b) Drag compensation.}
\label{fig:traj_jump}
\end{figure}

\begin{figure}[tbp]
\centerline{\includegraphics[width=0.5\textwidth]{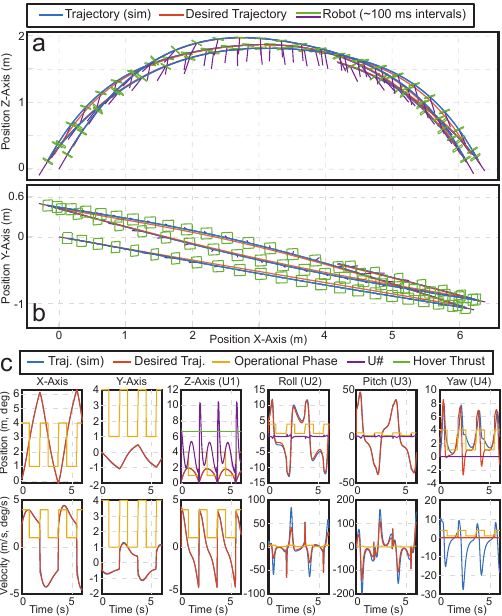}}
\caption{Hopping trajectory generation and control (Tables \ref{tab:hop_keyframes} and \ref{tab:traj_error} show details) for constant initial conditions (keyframe $\alpha_0$) including roll, pitch, and yaw of $[\phi, \theta, \psi] = [5,30,0]$ degrees, velocity magnitude aligned with the body z-axis of $5$ m/s, trajectory time $t_m = 1.75$ seconds (LO-TD), and reorientation time $\delta_t = 0.05$ seconds; where total simulation time equals $6$ seconds. The TD keyframe $\alpha_2$ for all generated trajectories includes $v_{TD_d}=5$ m/s and the desired orientation $[\phi, \theta, \psi]$ alternates between $[-5,-30,0]$ and $[5,30,0]$ degrees. The desired trajectories are drag compensated. a) Shows the z-x plane. b) Shows the y-x plane. c) Shows the individual states.}
\label{fig:traj_3D}
\end{figure}

Figures \ref{fig:traj_grd_wall} and \ref{fig:traj_jump} show the trajectory variability both with and without drag compensation (Table \ref{tab:traj_error}) for constant initial conditions (keyframe $\alpha_0$) including: roll, pitch, and yaw $[\phi,\theta, \psi]=[0,30,0]$ degrees, velocity magnitude of $5$ m/s aligned with the body z-axis, trajectory time $t_m = t_2 = 1.75$ seconds, reorientation time $t_1 = t_2-\delta_t$, and $\delta_t=50$ ms. Each incudes multiple trajectories generated over 6 seconds of operation; where Fig. \ref{fig:traj_3D} shows an example 3D trajectory with the individual states, operation phases, and inputs labeled. This variability from a constant $\alpha_0$ shows a potential for further increases in computational efficiency by precomputing the $P_l^{-1}$ and $N_l$ for a single or limited set of total trajectory times $t_m$; eliminating the required substitution of $t_m$ and $\delta_t$.

It has been shown that including linear drag in quadrotor trajectory generation can yield improvements in tracking performance \cite{svacha_improving_2017}. However, whereas quadrotors may be able to linearize drag about an operating point, high performance hopping robots necessarily undergo significant changes in velocity over the hop cycle; necessitating the use of non-linear drag (equations \ref{eq.d1}, \ref{eq.d2}). Table \ref{tab:traj_error} present a comparison between drag compensated trajectories (Figs. \ref{fig:traj_grd_wall}.b,d,f,h, \ref{fig:traj_jump}.b) and those without (Figs. \ref{fig:traj_grd_wall}.a,c,e,g, \ref{fig:traj_jump}.a); where removing the drag compensation requires setting $D_T=0$ in equation \ref{eq.4} and Table \ref{tab:hop_keyframes}. The RMSE in position and velocity show little difference for trajectories in Fig. \ref{fig:traj_grd_wall}.a-f where the TD keyframe $\alpha_2$ has free values as compared to the trajectories in Fig. \ref{fig:traj_grd_wall}.g,h with a full TD keyframe. Additionally, trajectories that maintain high horizontal velocity throughout, also show better performance when compensated for drag, as seen in Fig. \ref{fig:traj_jump}. Therefore, as expected, drag compensation has a bigger impact on the RMSE in position and velocity for more aggressive trajectories. Finally, as seen in comparing the trajectories in Fig. \ref{fig:traj_grd_wall}.a,b, free values in the TD keyframe $\alpha_2$ can yield very different trajectories (Table \ref{tab:hop_traj}). Therefore, if there is a generally preferred range of the free value (i.e., desire to jump-climb the wall or not), the null space $N_l$ can be used to shift the $c_j^*$ trajectory to achieve the desired result.

\begin{table}[tbp!]
\caption{Trajectory Error}
\begin{center}
\begin{tabular}{p{0.15\textwidth}|p{0.04\textwidth}p{0.05\textwidth}p{0.06\textwidth}p{0.07\textwidth}}
\hline
\textbf{Trajectory} & \textbf{Fig.} & \textbf{Drag} & \textbf{RMSE} & \textbf{RMSE} \\
\textbf{Type} & \textbf{Ref.} & \textbf{Comp.} & \textbf{Pos. (m)} & \textbf{Vel. (m/s)} \\
\hline
Wall-Wall (T2) & \ref{fig:traj_grd_wall}.a & NO & 0.072 & 0.243 \\
& \ref{fig:traj_grd_wall}.b & YES & 0.093 & 0.318 \\
\hline
Ground-Wall (T2) to& \ref{fig:traj_grd_wall}.c & NO & 0.099 & 0.287 \\
Wall-Wall (T2)& \ref{fig:traj_grd_wall}.d & YES & 0.090 & 0.304 \\
\hline
Wall-Ground (T1) to & \ref{fig:traj_grd_wall}.e & NO & 0.126 & 0.171\\
Ground-Ground (T1)& \ref{fig:traj_grd_wall}.f & YES & 0.119 & 0.160\\
\hline
Ground-State (T3)& \ref{fig:traj_grd_wall}.g & NO & 0.103& 0.222\\
& \ref{fig:traj_grd_wall}.h & YES & 0.034& 0.162\\
\hline
Ground-Ground (T1)& \ref{fig:traj_jump}.a & NO & 0.138& 0.295\\
& \ref{fig:traj_jump}.b & YES & 0.030& 0.152\\
\end{tabular}
\begin{tablenotes}
\item Error = desired (Pos./Vel.) - measured (Pos./Vel.). If two trajectory types are listed the first hop is the first listed, and the remaining hops are the second listed.
\end{tablenotes}
\label{tab:traj_error}
\end{center}
\end{table}

\section{Summary}
This work has presented a real-time, computationally efficient, non-linear drag compensated, trajectory generation methodology and accompanying Lyapunov-based controller for hopping robot locomotion. The methodology allows for the generation of trajectories from an initial keyframe (i.e. state) at liftoff to a final desired keyframe at touchdown. This includes those leaving from, and landing on, horizontal and vertical surfaces both with and without non-linear drag compensation. The presented methodology is broadly applicable to not only hopping robots but also quadrotors that desired greater control over their orientation while maintaining computational efficiency.

 

\bibliographystyle{IEEEtran.bst}
\bibliography{IEEEabrv,bibitems_v1}

\begin{thebibliography}{10}
\providecommand{\url}[1]{#1}
\csname url@samestyle\endcsname
\providecommand{\newblock}{\relax}
\providecommand{\bibinfo}[2]{#2}
\providecommand{\BIBentrySTDinterwordspacing}{\spaceskip=0pt\relax}
\providecommand{\BIBentryALTinterwordstretchfactor}{4}
\providecommand{\BIBentryALTinterwordspacing}{\spaceskip=\fontdimen2\font plus
\BIBentryALTinterwordstretchfactor\fontdimen3\font minus \fontdimen4\font\relax}
\providecommand{\BIBforeignlanguage}[2]{{%
\expandafter\ifx\csname l@#1\endcsname\relax
\typeout{** WARNING: IEEEtran.bst: No hyphenation pattern has been}%
\typeout{** loaded for the language `#1'. Using the pattern for}%
\typeout{** the default language instead.}%
\else
\language=\csname l@#1\endcsname
\fi
#2}}
\providecommand{\BIBdecl}{\relax}
\BIBdecl

\bibitem{burns_design_2025}
\BIBentryALTinterwordspacing
S.~Burns and M.~Woodward, ``\BIBforeignlanguage{en}{Design and {Control} of a {High}-{Performance} {Hopping} {Robot}},'' \emph{\BIBforeignlanguage{en}{IEEE Robotics and Automation Letters}}, vol.~10, no.~6, pp. 5641--5648, Jun. 2025.
\BIBentrySTDinterwordspacing

\bibitem{burns_optimized_2024}
\BIBentryALTinterwordspacing
S.~Burns and M.~Woodward, ``\BIBforeignlanguage{en}{Optimized {Kalman} {Filter} based {State} {Estimation} and {Height} {Control} in {Hopping} {Robots}},'' Nov. 2024, arXiv:2408.11978 [cs].
\BIBentrySTDinterwordspacing

\bibitem{bai_agile_2024}
S.~Bai, Q.~Pan, R.~Ding, H.~Jia, Z.~Yang, and P.~Chirarattananon, ``An agile monopedal hopping quadcopter with synergistic hybrid locomotion,'' \emph{Science Robotics}, vol.~9, no.~89, 2024.

\bibitem{wang_terrestrial_2023}
\BIBentryALTinterwordspacing
Y.~Wang, J.~Kang, Z.~Chen, and X.~Xiong, ``Terrestrial {Locomotion} of {PogoX}: {From} {Hardware} {Design} to {Energy} {Shaping} and {Step}-to-step {Dynamics} {Based} {Control},'' \emph{arXiv}, 2023, arXiv: 2309.13737.
\BIBentrySTDinterwordspacing

\bibitem{Raibert1984}
M.~H. Raibert, ``{Hopping in Legged Systems—Modeling and Simulation for the Two-Dimensional One-Legged Case},'' \emph{IEEE Transactions on Systems, Man and Cybernetics}, vol. SMC-14, no.~3, pp. 451--463, 1984.

\bibitem{Raibert1984a}
M.~H. Raibert, H.~B. Brown, and M.~Chepponis, ``{Experiments in Balance with a 3D One-Legged Hopping Machine},'' \emph{The International Journal of Robotics Research}, vol.~3, no.~2, pp. 75--92, 1984.

\bibitem{Zhu2022}
B.~Zhu, J.~Xu, A.~Charway, and D.~Saldana, ``{PogoDrone}: {Design}, {Model}, and {Control} of a {Jumping} {Quadrotor},'' \emph{Proceedings - IEEE International Conference on Robotics and Automation}, pp. 2031--2037, 2022, arXiv: 2204.00207 Publisher: IEEE ISBN: 9781728196817.

\bibitem{Haldane2016}
D.~W. Haldane, M.~M. Plecnik, J.~K. Yim, and R.~S. Fearing, ``Robotic vertical jumping agility via {Series}-{Elastic} power modulation,'' \emph{Science Robotics}, vol.~1, no.~1, 2016.

\bibitem{haldane_power_2016}
\BIBentryALTinterwordspacing
D.~W. Haldane, M.~Plecnik, J.~K. Yim, and R.~S. Fearing, ``\BIBforeignlanguage{en}{A power modulating leg mechanism for monopedal hopping},'' in \emph{\BIBforeignlanguage{en}{2016 {IEEE}/{RSJ} {International} {Conference} on {Intelligent} {Robots} and {Systems} ({IROS})}}.\hskip 1em plus 0.5em minus 0.4em\relax Daejeon, South Korea: IEEE, Oct. 2016, pp. 4757--4764.
\BIBentrySTDinterwordspacing

\bibitem{Plecnik2017}
M.~M. Plecnik, D.~W. Haldane, J.~K. Yim, and R.~S. Fearing, ``Design exploration and kinematic tuning of a power modulating jumping monopod,'' \emph{Journal of Mechanisms and Robotics}, vol.~9, no.~1, pp. 1--13, 2017.

\bibitem{lee_self-engaging_2018}
\BIBentryALTinterwordspacing
J.~S. Lee, M.~Plecnik, J.-H. Yang, and R.~S. Fearing, ``\BIBforeignlanguage{en}{Self-{Engaging} {Spined} {Gripper} with {Dynamic} {Penetration} and {Release} for {Steep} {Jumps}},'' in \emph{\BIBforeignlanguage{en}{2018 {IEEE} {International} {Conference} on {Robotics} and {Automation} ({ICRA})}}.\hskip 1em plus 0.5em minus 0.4em\relax Brisbane, QLD: IEEE, May 2018, pp. 1--8.
\BIBentrySTDinterwordspacing

\bibitem{haldane_repetitive_2017}
\BIBentryALTinterwordspacing
D.~W. Haldane, J.~K. Yim, and R.~S. Fearing, ``\BIBforeignlanguage{en}{Repetitive extreme-acceleration (14-g) spatial jumping with {Salto}-{1P}},'' in \emph{\BIBforeignlanguage{en}{2017 {IEEE}/{RSJ} {International} {Conference} on {Intelligent} {Robots} and {Systems} ({IROS})}}.\hskip 1em plus 0.5em minus 0.4em\relax Vancouver, BC: IEEE, Sep. 2017, pp. 3345--3351.
\BIBentrySTDinterwordspacing

\bibitem{yim_precision_2018}
\BIBentryALTinterwordspacing
J.~K. Yim and R.~S. Fearing, ``\BIBforeignlanguage{en}{Precision {Jumping} {Limits} from {Flight}-phase {Control} in {Salto}-{1P}},'' in \emph{\BIBforeignlanguage{en}{2018 {IEEE}/{RSJ} {International} {Conference} on {Intelligent} {Robots} and {Systems} ({IROS})}}.\hskip 1em plus 0.5em minus 0.4em\relax Madrid: IEEE, Oct. 2018, pp. 2229--2236.
\BIBentrySTDinterwordspacing

\bibitem{yim_drift-free_2019}
J.~K. Yim, E.~K. Wang, and R.~S. Fearing, ``Drift-free roll and pitch estimation for high-acceleration hopping,'' \emph{Proceedings - IEEE International Conference on Robotics and Automation}, vol. 2019-May, pp. 8986--8992, 2019, iSBN: 9781538660263.

\bibitem{Yim2020}
J.~K. Yim, B.~R.~P. Singh, E.~K. Wang, R.~Featherstone, and R.~S. Fearing, ``Precision robotic leaping and landing using stance-phase balance,'' \emph{IEEE Robotics and Automation Letters}, vol.~5, no.~2, pp. 3422--3429, 2020.

\bibitem{kang_fast_2024}
\BIBentryALTinterwordspacing
J.~Kang, Y.~Wang, and X.~Xiong, ``\BIBforeignlanguage{en}{Fast {Decentralized} {State} {Estimation} for {Legged} {Robot} {Locomotion} via {EKF} and {MHE}},'' May 2024, arXiv:2405.20567 [cs].
\BIBentrySTDinterwordspacing

\bibitem{hsiao_hybrid_2025}
H.~Hsiao, S.~Bai, Z.~Guan, S.~Kim, Z.~Ren, P.~Chirarattananon, and Y.~Chen, ``\BIBforeignlanguage{en}{Hybrid locomotion at the insect scale: {Combined} flying and jumping for enhanced efficiency and versatility},'' \emph{\BIBforeignlanguage{en}{Science Advances}}, 2025.

\bibitem{mellinger_minimum_2011}
\BIBentryALTinterwordspacing
D.~Mellinger and V.~Kumar, ``\BIBforeignlanguage{en}{Minimum {Snap} {Trajectory} {Generation} and {Control} for {Quadrotors}},'' in \emph{\BIBforeignlanguage{en}{2011 {IEEE} {International} {Conference} on {Robotics} and {Automation}}}.\hskip 1em plus 0.5em minus 0.4em\relax Shanghai, China: IEEE, May 2011, pp. 2520--2525.
\BIBentrySTDinterwordspacing

\bibitem{Bouabdallah2006}
S.~Bouabdallah, R.~Siegwart, and G.~Caprari, ``Design and control of an indoor coaxial helicopter,'' \emph{IEEE International Conference on Intelligent Robots and Systems}, no. April, pp. 2930--2935, 2006, iSBN: 142440259X.

\bibitem{svacha_improving_2017}
\BIBentryALTinterwordspacing
J.~Svacha, K.~Mohta, and V.~Kumar, ``\BIBforeignlanguage{en}{Improving quadrotor trajectory tracking by compensating for aerodynamic effects},'' in \emph{\BIBforeignlanguage{en}{2017 {International} {Conference} on {Unmanned} {Aircraft} {Systems} ({ICUAS})}}.\hskip 1em plus 0.5em minus 0.4em\relax Miami, FL, USA: IEEE, Jun. 2017, pp. 860--866.
\BIBentrySTDinterwordspacing

\bibitem{faessler_differential_2018}
\BIBentryALTinterwordspacing
M.~Faessler, A.~Franchi, and D.~Scaramuzza, ``\BIBforeignlanguage{en}{Differential {Flatness} of {Quadrotor} {Dynamics} {Subject} to {Rotor} {Drag} for {Accurate} {Tracking} of {High}-{Speed} {Trajectories}},'' \emph{\BIBforeignlanguage{en}{IEEE Robotics and Automation Letters}}, vol.~3, no.~2, pp. 620--626, Apr. 2018.
\BIBentrySTDinterwordspacing

\bibitem{van_nieuwstadt_realtime_1998}
M.~J. Van~Nieuwstadt and R.~M. Murray, ``Real‐time trajectory generation for differentially flat systems.pdf,'' \emph{International Journal Robust Nonlinear Control}, vol.~8, pp. 995--1020, 1998.

\end{thebibliography}

\newpage

\end{document}